\documentclass{article}

\usepackage[final, nonatbib]{neurips_2024}

\usepackage[utf8]{inputenc} 
\usepackage[T1]{fontenc}    
\usepackage{hyperref}       
\usepackage{url}            
\usepackage{booktabs}       
\usepackage{amsfonts}       
\usepackage{nicefrac}       
\usepackage{microtype}      
\usepackage{xcolor}         

\usepackage{multirow}
\usepackage[normalem]{ulem}
\useunder{\uline}{\ul}{}
\usepackage{graphicx}
\usepackage{amsmath} 
\usepackage{algorithm}
\usepackage{algpseudocode}
\usepackage{amsmath}
\usepackage{amsfonts}
\usepackage{bbm}
\usepackage{array}
\usepackage{subcaption}

\title{Tabular Data Generation using Binary Diffusion}

\author{%
  Vitaliy Kinakh \\
  Department of Computer Science \\
  University of Geneva \\
  Geneva, Switzerland \\
  \texttt{vitaliy.kinakh@unige.ch} \\
  \And
  Slava Voloshynovskiy \\
  Department of Computer Science \\
  University of Geneva \\
  Geneva, Switzerland \\
}

\begin{document}

\maketitle

\begin{abstract}

Generating synthetic tabular data is critical in machine learning, especially when real data is limited or sensitive. Traditional generative models often face challenges due to the unique characteristics of tabular data, such as mixed data types and varied distributions, and require complex preprocessing or large pretrained models. In this paper, we introduce a novel, lossless binary transformation method that converts any tabular data into fixed-size binary representations, and a corresponding new generative model called Binary Diffusion, specifically designed for binary data. Binary Diffusion leverages the simplicity of XOR operations for noise addition and removal and employs binary cross-entropy loss for training. Our approach eliminates the need for extensive preprocessing, complex noise parameter tuning, and pretraining on large datasets. We evaluate our model on several popular tabular benchmark datasets, demonstrating that Binary Diffusion outperforms existing state-of-the-art models on Travel, Adult Income, and Diabetes datasets while being significantly smaller in size. Code and models are available at: \url{https://github.com/vkinakh/binary-diffusion-tabular}

\end{abstract}

\section{Introduction}

The generation of synthetic tabular data is a critical task in machine learning, particularly when dealing with sensitive, private, or scarce real-world data. Traditional generative models often struggle with the inherent complexity and diversity of tabular data, which typically encompasses mixed data types and complex distributions.

In this paper, we introduce a method to transform generic tabular data into a binary representation, and a generative model named Binary Diffusion, specifically designed for binary data. Binary Diffusion leverages the simplicity of XOR operations for noise addition and removal, fundamental components of probabilistic diffusion models. This method eliminates the need for extensive preprocessing and complex noise parameter tuning, streamlining the data preparation process.


Our approach offers several key advantages. First, by converting all columns into unified binary representations, the proposed transformation removes the necessity for column-specific preprocessing commonly required in handling mixed-type tabular data. Secondly, the Binary Diffusion model itself is optimized for binary data, utilizing binary cross-entropy (BCE) loss for predictions during the training of the denoising network.

We evaluate our model on several popular tabular benchmark datasets, including Travel \cite{tejashvi2021tour}, Sick \cite{smith1988using}, HELOC \cite{liabev2018heloc, fico2018challenge}, Adult Income \cite{adult_2}, California Housing \cite{pace1997sparse, nugent2017california}, and Diabetes \cite{strack2014impact, KaggleDiabetesReadmission2021} tabular datasets. The Binary Diffusion model outperforms existing state-of-the-art models on Travel, Adult Income and Dianetes datasets. Additionally, our model is significantly smaller in size compared to contemporary models and does not require pretraining on other data modalities, unlike methods based on large language models (LLMs) such as GReaT \cite{borisov2022language}.


\section{Related Work}

\textbf{TVAE} (Tabular Variational Autoencoder) adapts the Variational Autoencoder (VAE) framework to handle mixed-type tabular data by separately modeling continuous and categorical variables. \textbf{CTGAN} (Conditional Tabular GAN) employs a conditional generator to address imbalanced categorical columns, ensuring the generation of diverse and realistic samples by conditioning on categorical data distributions. \textbf{CopulaGAN} integrates copulas with GANs to capture dependencies between variables, ensuring that synthetic data preserves the complex relationships present in the original dataset \cite{xu2019modeling}.

\textbf{GReaT} (Generation of Realistic Tabular data) \cite{borisov2022language} leverages a pretrained auto-regressive language model (LLM) to generate highly realistic synthetic tabular data. The process involves fine-tuning the LLM on textually encoded tabular data, transforming each row into a sequence of words. This approach allows the model to condition on any subset of features and generate the remaining data without additional overhead.

Existing data generation methods show several shortcomings. Models such as CopulaGAN, CTGAN, and TVAE attempt to generate columns with both continuous and categorical data simultaneously, employing activation functions like {\it softmax} and {\it tanh} in the outputs. These models also require complex preprocessing of continuous values and rely on restrictive approximations using Gaussian mixture models and mode-specific normalization. Additionally, large language model-based generators like GReaT need extensive pretraining on text data, making them computationally intensive with large parameter counts with potential bias from the pretraining data.


The proposed data transformation and generative model address these shortcomings as follows: (i) by converting all columns to unified binary representations; (ii) the proposed generative model for binary data, with fewer than 2M parameters, does not require pretraining on large datasets and offers both fast training and sampling capabilities.

\section{Data transformation}

\begin{figure}[htbp]
    \centering
    \includegraphics[width=0.9\textwidth]{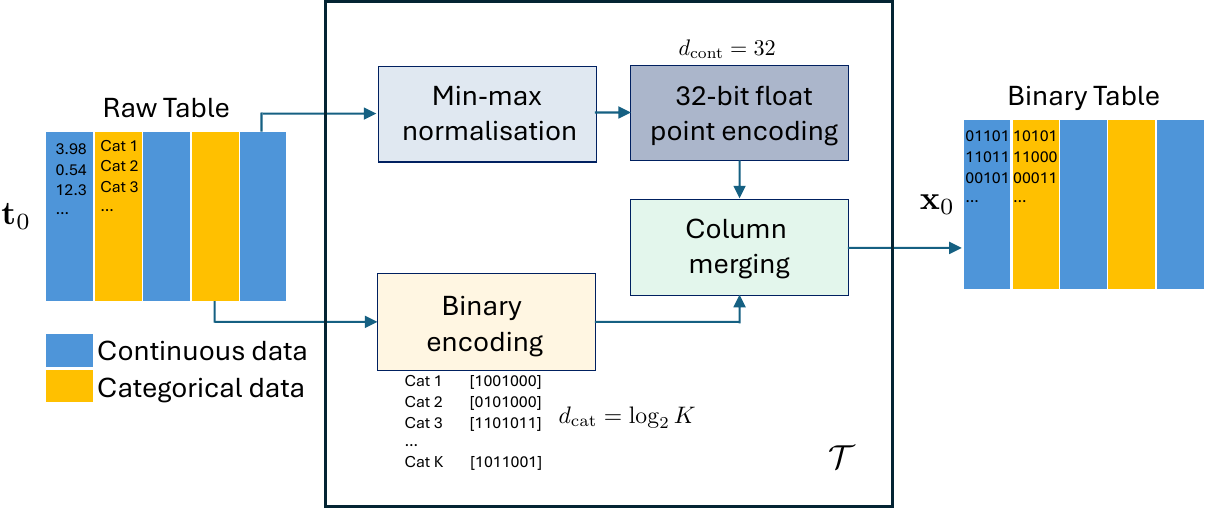}
    \caption{Transformation of tabular data ${\bf t}_0$ into the binary form ${\bf x}_0$. The considered transformation is reversible. The continuous column records are presented with the length $d_{\text {cont }}=32$ and the categorical ones with $d_{\text {cat }}=\log _2 K$, where $K$ stands for the number of categorical classes.}
    \label{fig:transformation}
\end{figure}

 To apply the Binary Diffusion model to tabular data, we propose an invertible lossless transformation $\mathcal{T}$, shown on the Figure \ref{fig:transformation}, that converts tabular data columns into fixed-size binary representations. The transformations is essential for preparing tabular data for the Binary Diffusion model, enabling it to process and generate tabular data without the need for extensive preprocessing. This approach ensures that the data retains its original characteristics.
 
The transformation method converts each column of the table into a binary format. For continuous data, this process includes applying min-max normalization to the columns, followed by converting these normalized values into a binary representation via 32-bit floating-point encoding. For categorical data, binary encoding is used. The encoded columns are concatenated into fixed-size rows. 

The inverse transformation $\mathcal{T}^{-1}$  converts the binary representations back into their original form. For continuous data, the decoded values are rescaled to their original range using metadata generated during the initial transformation. For categorical data, the binary codes are mapped back to their respective categories using a predefined mapping scheme. 

\section{Binary Diffusion}

\begin{figure}[htbp]
    \centering
    \includegraphics[width=1.0\textwidth]{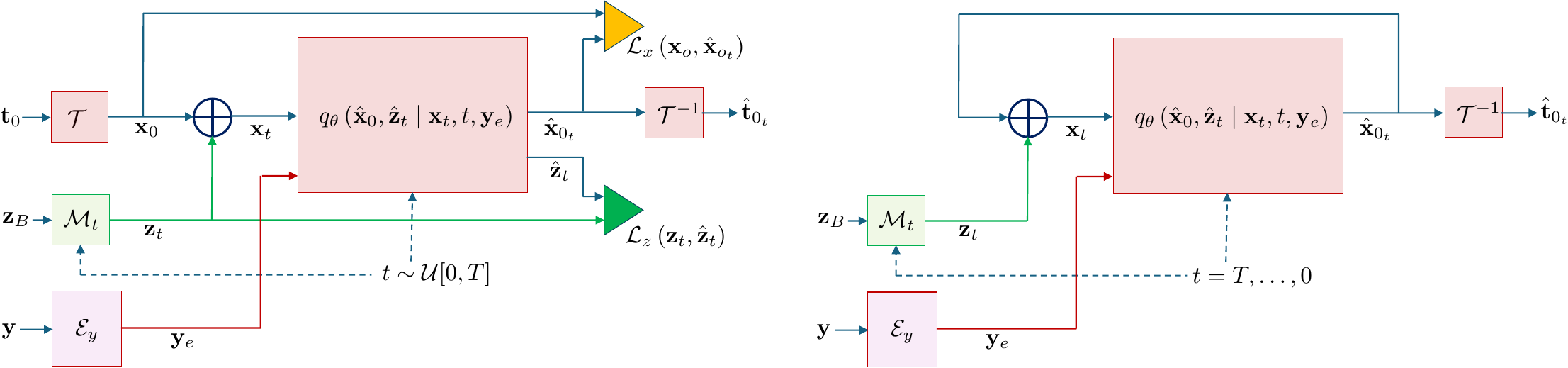}
    \caption{Binary Diffusion training (left) and sampling (right) schemes.}
    \label{fig:training_sampling}
\end{figure}

Binary Diffusion shown in Figure \ref{fig:training_sampling} is a novel approach for generative modeling that leverages the simplicity and robustness of binary data representations. This method involves adding and removing noise through XOR operation, which makes it particularly well-suited for handling binary data. Below, we describe the key aspects of the Binary Diffusion methodology in detail.

In Binary Diffusion, noise is added to the data by flipping bits using the XOR operation with a random binary mask. The amount of noise added is quantified by the proportion of bits flipped. Let $\textbf{x}_0 \in \{0, 1\}^d$ be the original binary vector of dimension $d$, and $\textbf{z}_t \in \{0, 1\}^d$ be a random binary noise vector at timestep $t$. The noisy vector $\textbf{x}_t$ is obtained as: $\textbf{x}_t = \textbf{x}_0 \oplus \textbf{z}_t$ , where $\oplus$ denotes the XOR operation. The noise level is defined as the fraction of bits flipped in $\textbf{z}_t$ in the mapper $\mathcal{M}_t$ at step $t$, with the number of bits flipped ranging within $[0, 0.5]$ as a function of the timestep.

The denoising network $q_\theta(\hat{\textbf{x}}_0, \hat{\textbf{z}}_t | \textbf{x}_t, t, \textbf{y}_e)$ is trained to predict both the added noise $\textbf{z}_t$ and the clean-denoised vector $\textbf{x}_0$ from the noisy vector $\textbf{x}_t$. We employ binary cross-entropy (BCE) loss (\ref{eq:binary_diffusion_loss}) to train the denoising network. The loss function is averaged over both the batch of samples and the dimensions of the vectors:

\begin{equation}
\scalebox{0.95}{$
\label{eq:binary_diffusion_loss}
\begin{aligned}
\mathcal{L}(\theta) &= \frac{1}{B} \sum_{b=1}^B \left[ \mathcal{L}_{x}(\hat{\textbf{x}}_0^{(b)}, \textbf{x}_0^{(b)}) + \mathcal{L}_{z}(\hat{\textbf{z}}_t^{(b)}, \textbf{z}_t^{(b)}) \right] \\
&= -\frac{1}{B} \sum_{b=1}^B \sum_{i=1}^d \left[ \textbf{x}_{0i}^{(b)} \log \hat{\textbf{x}}_{0i}^{(b)} + (1 - \textbf{x}_{0i}^{(b)}) \log (1 - \hat{\textbf{x}}_{0i}^{(b)}) \right] \\
&\quad - \frac{1}{B} \sum_{b=1}^B \sum_{i=1}^d \left[ \textbf{z}_{ti}^{(b)} \log \hat{\textbf{z}}_{ti}^{(b)} + (1 - \textbf{z}_{ti}^{(b)}) \log (1 - \hat{\textbf{z}}_{ti}^{(b)}) \right],
\end{aligned}
$}
\end{equation}
where $B$ is the batch size, $\theta$ represents the parameters of the denoising network,  $\textbf{x}_0^{(b)}$ and $\hat{\textbf{x}}_0^{(b)}$ are the $b$ -th samples of the true clean vectors and the predicted clean vectors, respectively. Similarly, $\textbf{z}_t^{(b)}$ and $\hat{\textbf{z}}_t^{(b)}$ are the $b$-th samples of the true added noise vectors and the predicted noise vectors, respectively. $\textbf{y}_e = \mathcal{E}_y({\bf y})$ denotes the encoded label $\bf y$, one-hot encoded for classification and min-max normalized for regression. $\mathcal{L}_{x}$ and $\mathcal{L}_{z}$ denotes binary cross-entropy (BCE) loss. The indices $i$ and $b$ correspond to the $i$-th dimension of the vectors and the $b$-th sample in the batch, respectively.

During training (Figure \ref{fig:training_sampling} left), we use classifier-free guidance \cite{ho2022classifier}. For classification tasks, the conditioning input class label $\textbf{y}$ is a one-hot encoded label $\textbf{y}_e$. For regression tasks, $\textbf{y}$ consists of min-max normalized target values $\textbf{y}_e$, allowing the model to generate data conditioned on specific numerical outcomes. In unconditional training, we use an all-zeros conditioning vector for classification tasks and a value of $-1$ for regression tasks to indicate the absence of conditioning.

When sampling (Figure \ref{fig:training_sampling} right), we start from a random binary vector $\textbf{x}_t$ at timestep $t = T$, along with the conditioning variable $\textbf{y}$, encoded into $\textbf{y}_e$. For each selected timestep in the sequence $[T, \ldots, 0]$, denoising is applied to the vector. The denoised vector $\hat{\textbf{x}}_0$ and the estimated binary noise $\hat{\textbf{z}}_t$ are predicted by the denoising network. These predictions are then processed using a sigmoid function and binarized with a threshold. During sampling, we use the denoised vector  $\hat{\textbf{x}}_0$ directly. Then, random noise $\textbf{z}_t$ is generated and added to $\hat{\textbf{x}}_0$ via the XOR operation: $\textbf{x}_t = \hat{\textbf{x}}_0 \oplus \textbf{z}_t$. The sampling algorithm is summarized in Algorithm \ref{alg:sampling}.

\section{Results}
\label{sec:results}

We evaluate the performance of Binary Diffusion on widely-recognized tabular benchmark datasets, including Travel \cite{tejashvi2021tour}, Sick \cite{smith1988using}, HELOC \cite{liabev2018heloc, fico2018challenge}, Adult Income \cite{adult_2}, California Housing \cite{pace1997sparse, nugent2017california}, and Diabetes \cite{strack2014impact, KaggleDiabetesReadmission2021}. For classification tasks (Travel, Sick, HELOC, Adult Income, and Diabetes), classification accuracy was used as metric, while mean squared error (MSE) was used for the regression task (California Housing). Following the evaluation protocol established in \cite{borisov2022language}, we employed Linear/Logistic Regression (LR), Decision Tree (DT), and Random Forest (RF) as downstream models to assess the quality of the synthetic data. The datasets were split into training and test sets with an 80/20 split. The generative models were trained on the training set, and the test set was reserved for evaluation. To ensure robustness, 5 sets of synthetic training data were generated, and the results are reported as average performances with corresponding standard deviations. Table \ref{table:results} shows the detailed results. Binary Diffusion achieved superior performance compared to existing state-of-the-art models on the Travel, Adult Income, and Diabetes datasets. Notably, Binary Diffusion maintained competitive results on the HELOC and Sick datasets, despite having a significantly smaller parameter footprint (ranging from 1.1M to 2.6M parameters) compared to models like GReaT, which utilize large language models with hundreds of millions of parameters. Binary Diffusion does not require pretraining on external data modalities, enhancing its efficiency and reducing potential biases associated with pretraining data. In the regression task (California Housing), Binary Diffusion demonstrated competitive MSE scores. Additionally, Binary Diffusion offers faster training and sampling times, as detailed in Appendix \ref{sec:runtime_comparison}. Implementation details are summarized in Appendix \ref{sec:implementation_details}.

\begin{table}[h]
\centering
\caption{Quantitative results on table dataset benchmarks. The best results are marked in \textbf{bold}, second-best are {\ul underlined}. The number of parameters for every model and dataset are provided in $4$-th row for every dataset. }
\label{table:results}

\resizebox{1.0\textwidth}{!}{

\begin{tabular}{lcccccccc}
\hline
Dataset                     & Model & Original   & TVAE        & CopulaGAN  & CTGAN      & Distill-GReaT  & GReaT            & Binary Diffusion \\ \hline
\multirow{4}{*}{Travel ($\uparrow$)} & LR    & 82.72$\pm$0.00 & 79.58$\pm$0.00  & 73.30$\pm$0.00 & 73.30$\pm$0.00 & 78.53$\pm$0.00     & {\ul 80.10$\pm$0.00} & \textbf{83.79$\pm$0.08} \\
                            & DT    & 89.01$\pm$0.00 & 81.68$\pm$1.28  & 73.61$\pm$0.26 & 73.30$\pm$0.00 & 77.38$\pm$0.51     & {\ul 83.56$\pm$0.42} & \textbf{88.90$\pm$0.57} \\
                            & RF    & 85.03$\pm$0.53 & 81.68$\pm$1.19  & 73.30$\pm$0.00 & 71.41$\pm$0.53 & 79.50$\pm$0.53     & {\ul 84.30$\pm$0.33} & \textbf{89.95$\pm$0.44} \\
                             & Params &  - & 36K & 157K & 155K & 82M & 355M & 1.1M \\
                            \hline
\multirow{4}{*}{Sick ($\uparrow$)}   & LR    & 96.69$\pm$0.00 & 94.70$\pm$0.00  & 94.57$\pm$0.00 & 94.44$\pm$0.00 & {\ul 96.56$\pm$0.00} & \textbf{97.72$\pm$0.00} & 96.14$\pm$0.63        \\
                            & DT    & 98.94$\pm$0.12 & 95.39$\pm$0.18 & 93.77$\pm$0.01 & 92.05$\pm$0.41 & 95.39$\pm$0.18 & \textbf{97.72$\pm$0.00} & {\ul 97.07$\pm$0.24}          \\
                            & RF    & 98.28$\pm$0.06 & 94.91$\pm$0.06  & 94.57$\pm$0.01 & 94.57$\pm$0.00 & {\ul 97.72$\pm$0.00} & \textbf{98.30$\pm$0.13} & 96.59$\pm$0.55          \\
                            & Params & - & 46K & 226K & 222K & 82M & 355M & 1.4M \\
                            \hline
\multirow{4}{*}{HELOC ($\uparrow$)}  & LR    & 71.80$\pm$0.00 & 71.04$\pm$0.00  & 42.03$\pm$0.00 & 57.72$\pm$0.00 & 70.58$\pm$0.00      & \textbf{71.90$\pm$0.00} & {\ul 71.76$\pm$0.30}    \\
                            & DT    & 81.90$\pm$1.06 & {\ul 76.39$\pm$0.50} & 42.36$\pm$0.10 & 61.34$\pm$0.09 & 81.40$\pm$0.15  & \textbf{79.10$\pm$0.07} & 70.25$\pm$0.43          \\
                            & RF    & 83.19$\pm$0.71 & 77.24$\pm$0.25  & 42.35$\pm$0.34 & 62.35$\pm$0.35 & \textbf{82.14$\pm$0.13} & {\ul 80.93$\pm$0.28} & 70.47$\pm$0.32          \\
                            & Params & - & 62K & 276K & 277K & 82M & 355M & 2.6M \\
                            \hline
\multirow{4}{*}{Adult Income ($\uparrow$)} & LR & 85.00$\pm$0.00 & 80.53$\pm$0.00 & 80.61$\pm$0.00 & 83.20$\pm$0.00 & 84.65$\pm$0.00 & {\ul 84.77$\pm$0.00} & \textbf{85.45$\pm$0.11} \\
                                  & DT & 85.27$\pm$0.01 & 82.80$\pm$0.08 & 76.29$\pm$0.06 & 81.32$\pm$0.02 & 84.49$\pm$0.04 & {\ul 84.81$\pm$0.04} & \textbf{85.27$\pm$0.11} \\
                                  & RF & 85.93$\pm$0.11 & 83.48$\pm$0.11 & 80.46$\pm$0.21 & 83.53$\pm$0.07 & 85.25$\pm$0.07 & {\ul 85.42$\pm$0.05} & \textbf{85.74$\pm$0.11}           \\
                                  & Params & - & 53K & 300K & 302K & 82M & 355M & 1.4M \\
                                  \hline
\multirow{4}{*}{Diabetes ($\uparrow$)} & LR & 58.76$\pm$0.00 & 56.34$\pm$0.00 & 40.27$\pm$0.00 & 50.93$\pm$0.00 & 57.33$\pm$0.00 & 57.34$\pm$0.00 & \textbf{57.75$\pm$0.04} \\
                                  & DT & 57.29$\pm$0.03 & 53.30$\pm$0.08 & 38.50$\pm$0.02 & 49.73$\pm$0.02 & 54.10$\pm$0.04 & 55.23$\pm$0.04 & \textbf{57.13$\pm$0.15} \\
                                  & RF & 59.00$\pm$0.08 & 55.17$\pm$0.10 & 37.59$\pm$0.31 & 52.23$\pm$0.17 & {\ul 58.03$\pm$0.16} & \textbf{58.34$\pm$0.09} & 57.52$\pm$0.12           \\
                                  & Params & - & 369K & 9.4M & 9.6M & 82M & 355M & 1.8M \\
                                  \hline
                                  
\multirow{4}{*}{California Housing ($\downarrow$)} & LR    & 0.40$\pm$0.00  & 0.65$\pm$0.00        & 0.98$\pm$0.00  & 0.61$\pm$0.00  & 0.57$\pm$0.00           & \textbf{0.34$\pm$0.00}           &  {\ul 0.55$\pm$0.00} \\
                                       & DT    & 0.32$\pm$0.01  & 0.45$\pm$0.01        & 1.19$\pm$0.01  & 0.82$\pm$0.01  & {\ul 0.43$\pm$0.01}           & \textbf{0.39$\pm$0.01}           &  0.45$\pm$0.00 \\
                                       & RF    & 0.21$\pm$0.01  & 0.35$\pm$0.01        & 0.99$\pm$0.01  & 0.62$\pm$0.01  & {\ul 0.32$\pm$0.01}           & \textbf{0.28$\pm$0.01}           &   0.39$\pm$0.00  \\
                                       & Params & - & 45K & 201K & 197K & 82M & 355M & 1.5M \\
                                       \hline
\end{tabular}
}

\end{table}

\section{Conclusions}

This paper proposed a novel lossless binary transformation method for tabular data, which converts any data into fixed-size binary representations. Building upon this transformation, we introduced the Binary Diffusion model, a generative model specifically designed for binary data that utilizes XOR operations for noise addition and removal and is trained using binary cross-entropy loss. Our approach addresses several shortcomings of existing methods, such as the need for complex preprocessing, reliance on large pretrained models, and computational inefficiency.

We evaluated our model on several tabular benchmark datasets, and demonstrated that Binary Diffusion achieves state-of-the-art performance on these datasets while being significantly smaller in size compared to existing models. Our model does not require pretraining on other data modalities, which simplifies the training process and avoids potential biases from pretraining data. Our findings indicate that the proposed model works particularly well with datasets that have a high proportion of categorical columns.


{
    \small
    \bibliographystyle{alpha}
    \bibliography{main}
}


\newpage
\appendix

\section{Sampling algorithm}
\label{sec:sampling_algorithm}

\begin{algorithm}
\caption{Sampling algorithm.}
\label{alg:sampling}
\begin{algorithmic}[1]
\State $\textbf{x}_t \gets$ random binary tensor
\State $\textbf{y} \gets$ condition/label
\State $\textbf{y}_e \gets$ apply condition enxoding 
\State $threshold \gets$ threshold value to binarize \Comment{Default 0.5}
\State $q_\theta(\hat{\textbf{x}}_0, \hat{\textbf{z}}_t | \textbf{x}_t, t, \textbf{y}_e) \gets$ pre-trained denoiser network

\For{$t \in \{T, \dots, 0\}$} \Comment{Selected timesteps}
    \State $\hat{\textbf{x}}_0, \hat{\textbf{z}}_t \gets q_\theta(\hat{\textbf{x}}_0, \hat{\textbf{z}}_t | \textbf{x}_t, t, \textbf{y}_e)$
    \State $\hat{\textbf{x}}_0 \gets \sigma(\hat{\textbf{x}}_0) > threshold$ \Comment{Apply sigmoid and compare to threshold}

    \State $\textbf{z}_t \gets get\_binary\_noise(t)$ \Comment{Generate random noise}
    \State $\textbf{x}_t \gets \hat{\textbf{x}}_0 \oplus \textbf{z}_t$ \Comment{Update $\textbf{x}_t$ using XOR with $\textbf{z}_t$}
\EndFor

\State \textbf{return} $\textbf{x}_t$
\end{algorithmic}
\end{algorithm}

\section{Evaluation models hyperparameters}
\label{sec:eval_models_hyperparameters}

During evaluation, we follow the evaluation proposed in \cite{borisov2022language}. The hyperparameter configuration of the evaluation models for the ML efficiency experiments are provided in Table \ref{table:eval_models_hyperparameters}.

\begin{table}[hb!]
\centering
\caption{Evaluation models hyperparameters.}
\label{table:eval_models_hyperparameters}
\begin{tabular}{lcccc}
\hline
                   & LR        & DT         & \multicolumn{2}{c}{RF}     \\ \cline{2-5} 
Dataset            & max\_iter & max\_depth & max\_depth & n\_estimators \\ \hline
Travel             & 100       & 6          & 12         & 75            \\
Sick               & 200       & 10         & 12         & 90            \\
HELOC              & 500       & 6          & 12         & 78            \\
Adult Income       & 1000      & 8          & 12         & 85            \\
Diabetes           & 500       & 10         & 20         & 120           \\
California Housing & -         & 10         & 12         & 85            \\ \hline
\end{tabular}
\end{table}

\section{Runtime comparison}
\label{sec:runtime_comparison}

We compare the training and sampling times, the number of training epochs, batch sizes, and peak VRAM utilization of generative models. The results, including the number of training epochs and batch sizes required for each model to converge, are summarized in Table \ref{table:times}. Specifically, for TVAE, CopulaGAN, and CTGAN, we employed the default batch size of 500 and trained for 200 epochs; for Distill-GReaT and GReaT, we used a batch size of 32 and trained for 200 epochs; and for Binary Diffusion, a batch size of 256 and 500 epochs were utilized to ensure model convergence. For this study, we utilized the Adult Income dataset. All experiments were conducted on a PC with a single RTX 2080 Ti GPU, an Intel Core i9-9900K CPU 3.60 GHz with 16 threads, 64 GB of RAM, and Ubuntu 20.04 LTS as the operating system.

\begin{table}[h]
\centering
\caption{Comparison of training and sampling times, and peak VRAM utilization.}
\label{table:times}

\resizebox{\textwidth}{!}{
\begin{tabular}{lccccc}
\hline
Model            &  Epochs & Batch size & Training time & Sampling time (s) & Peak VRAM use \\ \hline
TVAE             & 200             & 500            & 2 min 21 sec  & 0.036 $\pm$ 0.001     & 240 MiB      \\
CopulaGAN        & 200             & 500            & 4 min 26 sec  & 0.101 $\pm$ 0.003    & 258 MiB      \\
CTGAN            & 200             & 500            & 4 min 33 sec  & 0.055 $\pm$ 0.005    & 258 MiB      \\
Distill-GReaT    & 200             & 32             & 5 h 7 min     & 7.104 $\pm$ 0.025    & 8184 MiB     \\
GReaT            & 200             & 32             & 7 h 33 min    & 11.441 $\pm$ 0.034   & 8548 MiB     \\
Binary Diffusion & 5000            & 256            & 53 min 2 sec  & 0.347 $\pm$ 0.006    & 266 MiB      \\ \hline
\end{tabular}
}
\end{table}

\section{Implementation details}
\label{sec:implementation_details}

\textbf{Denoiser Architecture}. We use a similar denoiser architecture across all datasets, which takes as input a noisy vector $x_t$ of size $d$, a timestep $t$, and an input condition $y$. The input size $d$ corresponds to the size of the binary vector in each dataset. The input vector $x_t$ is projected through a linear layer with 256 output units. The timestep $t$ is processed using a sinusoidal positional embedding, followed by two linear layers with 256 output units each, interleaved with GELU activation functions \cite{hendrycks2016gaussian}. The input condition $y$ is processed through a linear projector with 256 output units. The outputs of the timestep embedding and the condition projector are then combined via element-wise addition. This combined representation is subsequently processed by three ResNet blocks that incorporate timestep embeddings. Depending on the size of the binary representation for each dataset, the number of parameters varies between 1.1 million and 1.4 million.

\textbf{Training and Sampling Details}. We trained the denoiser for 50,000 steps using the Adam optimizer \cite{kingma2014adam} with a learning rate of $1 \times 10^{-4}$, a weight decay of 0, and a batch size of 256. To maintain a distilled version of the denoiser, we employed an Exponential Moving Average (EMA) with a decay rate of 0.995, updating it every 10 training steps. This distilled model was subsequently used for sampling. During training, we utilized classifier-free guidance with a 10\% probability of using a zero token. The diffusion model was configured to perform 1,000 denoising steps during training. Given the relatively small size of our models, we opted for full-precision training. All training parameters are summarized in Table \ref{table:training_details}.

\begin{table}[h]
\centering
\caption{Binary Diffusion training details.}
\label{table:training_details}
\begin{tabular}{l|l}
config                              & value    \\ \hline
optimizer                           & Adam     \\
learning rate                       & 1e-4 \\
weight decay                        & 0        \\
batch size                          & 256      \\
training steps                      & 500000   \\
EMA decay                           & 0.995    \\
EMA update frequency                & 10       \\
classifier-free guidance zero token & 0.1      \\
precision                           & fp32     \\
diffusion timesteps                 & 1000
\end{tabular}
\end{table}

We empirically observed that model performance, measured by accuracy for classification tasks and mean squared error (MSE) for regression tasks deteriorates as the number of sampling steps increases. We selected 5 sampling steps and a guidance scale of 5 for all datasets to optimize performance.

\begin{table}[h]
\centering
\caption{Binary Diffusion sampling details.}
\label{table:sampling_details}
\begin{tabular}{l|l}
config         & value \\ \hline
sampling steps & 5     \\
guidance scale & 5     \\
EMA            & True 
\end{tabular}
\end{table}

\textbf{Environment}. All experiments were conducted on a PC with a single RTX 2080 Ti GPU, an Intel Core i9-9900K CPU 3.60 GHz with 16 threads, 64 GB of RAM, and Ubuntu 20.04 LTS as the operating system. We utilized PyTorch \cite{paszke2019pytorch} with the Accelerate \cite{accelerate} library for training generative models, and the scikit-learn \cite{scikit_learn} library for evaluating models.

\section{Effect of sampling steps}
\label{sec:sampling_steps}

\begin{figure}[ht]
    \centering
    \begin{subfigure}[b]{0.4\textwidth}
        \centering
        \includegraphics[width=\linewidth]{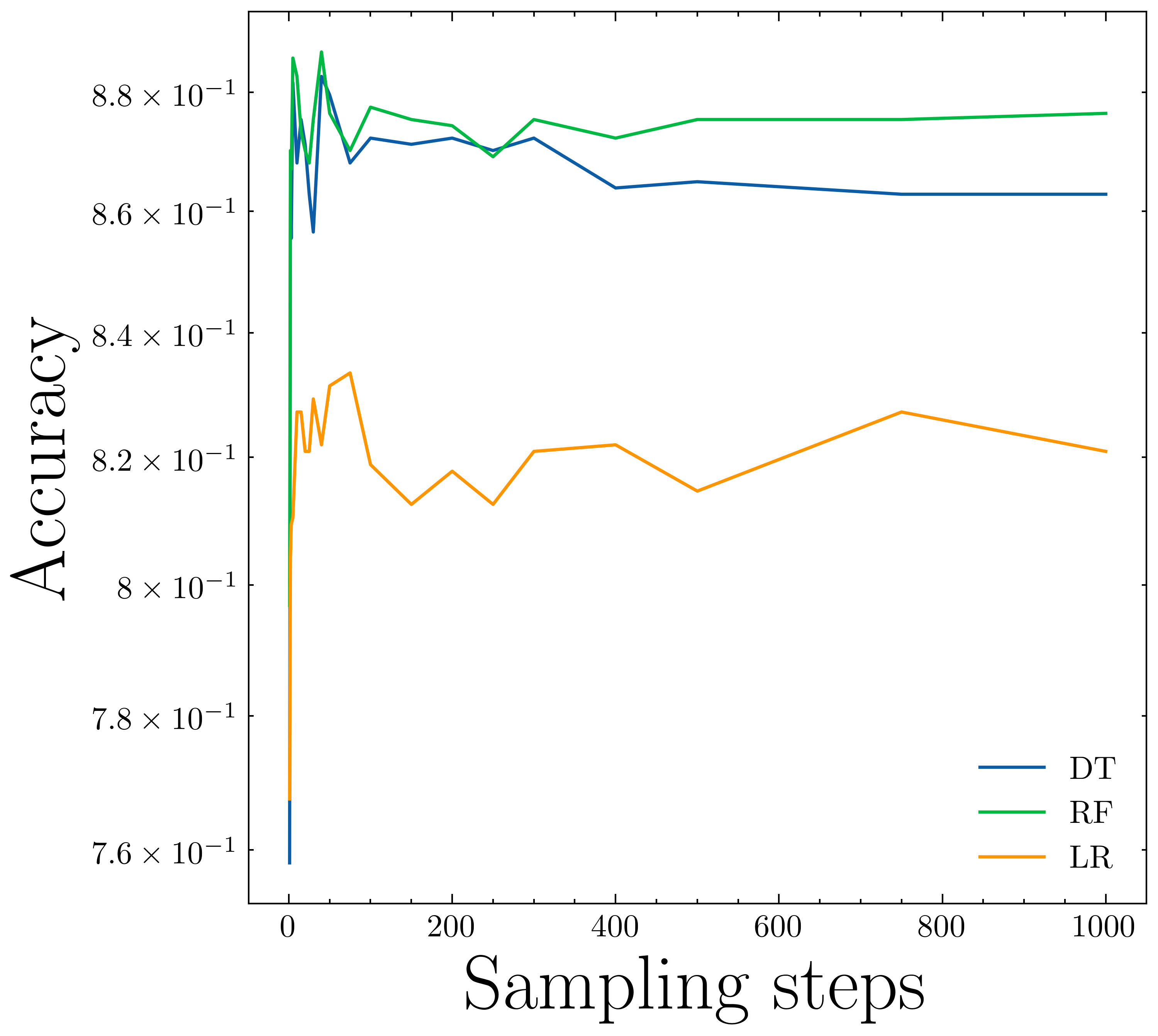}
        \caption{Travel}
        \label{fig:travel_steps}
    \end{subfigure}
    \hfill
    \begin{subfigure}[b]{0.4\textwidth}
        \centering
        \includegraphics[width=\linewidth]{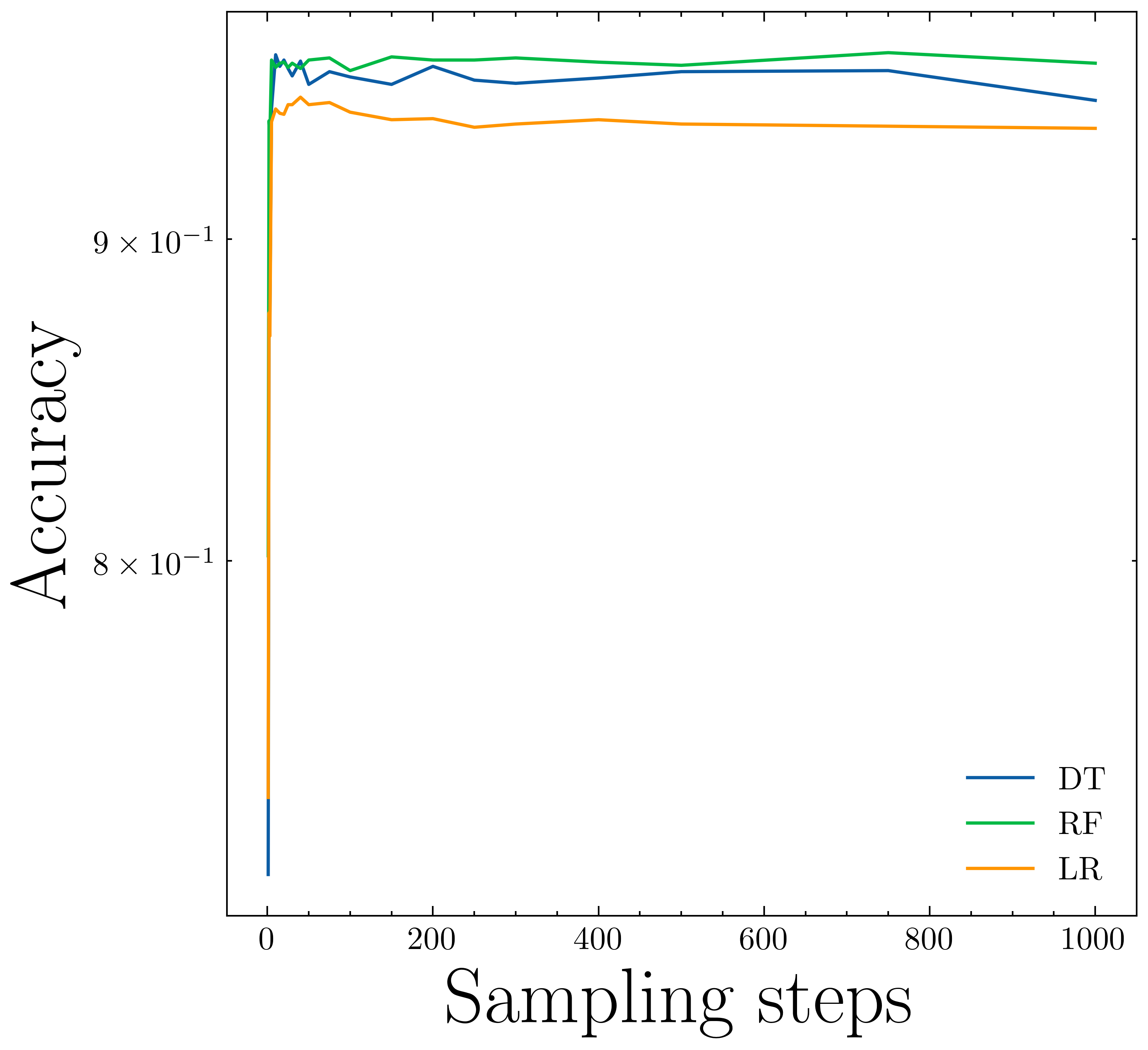}
        \caption{Sick}
        \label{fig:sick_steps}
    \end{subfigure}
    
    \vspace{0.5cm}
    
    \begin{subfigure}[b]{0.4\textwidth}
        \centering
        \includegraphics[width=\linewidth]{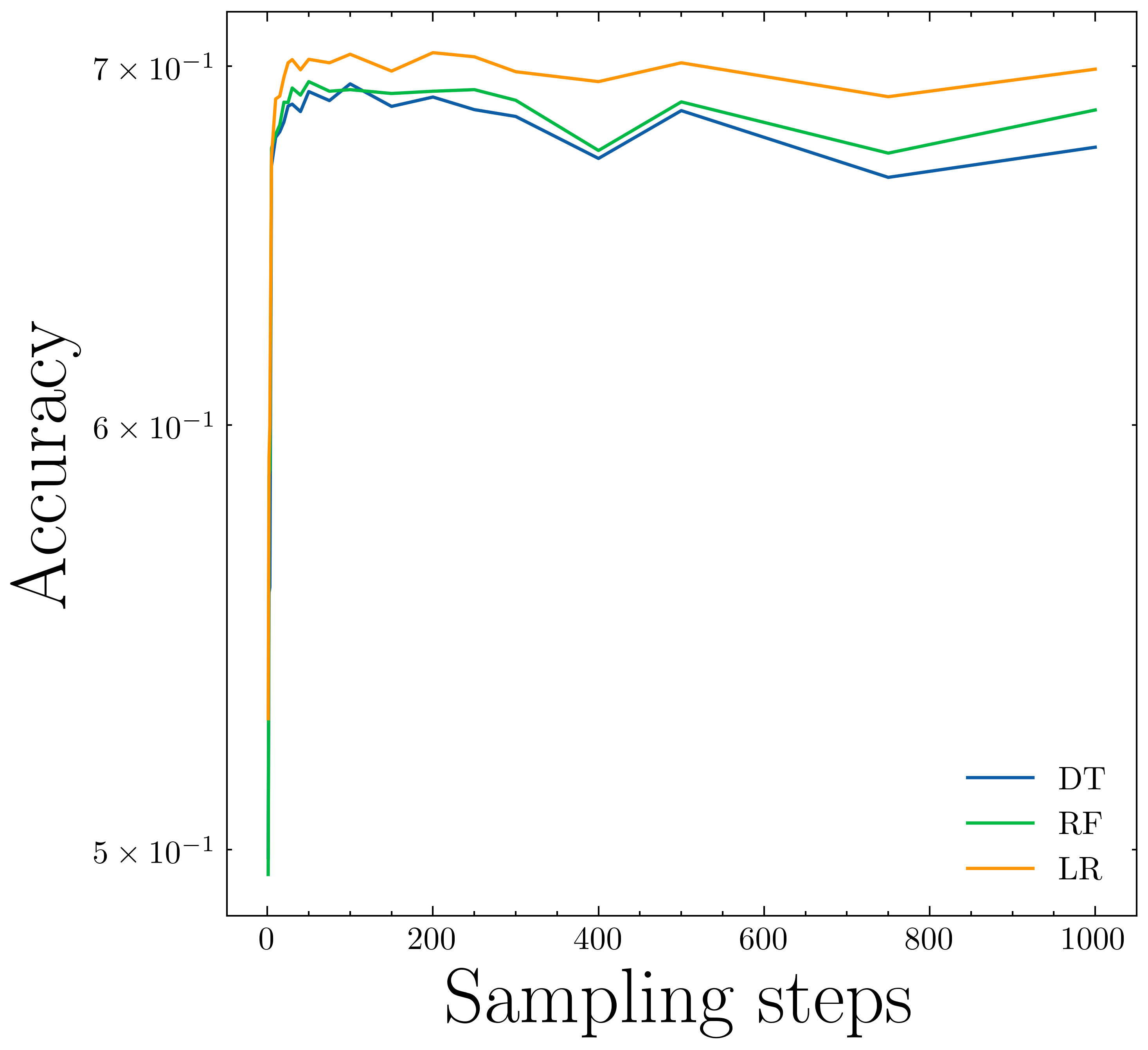}
        \caption{HELOC}
        \label{fig:heloc_steps}
    \end{subfigure}
    \hfill
    \begin{subfigure}[b]{0.4\textwidth}
        \centering
        \includegraphics[width=\linewidth]{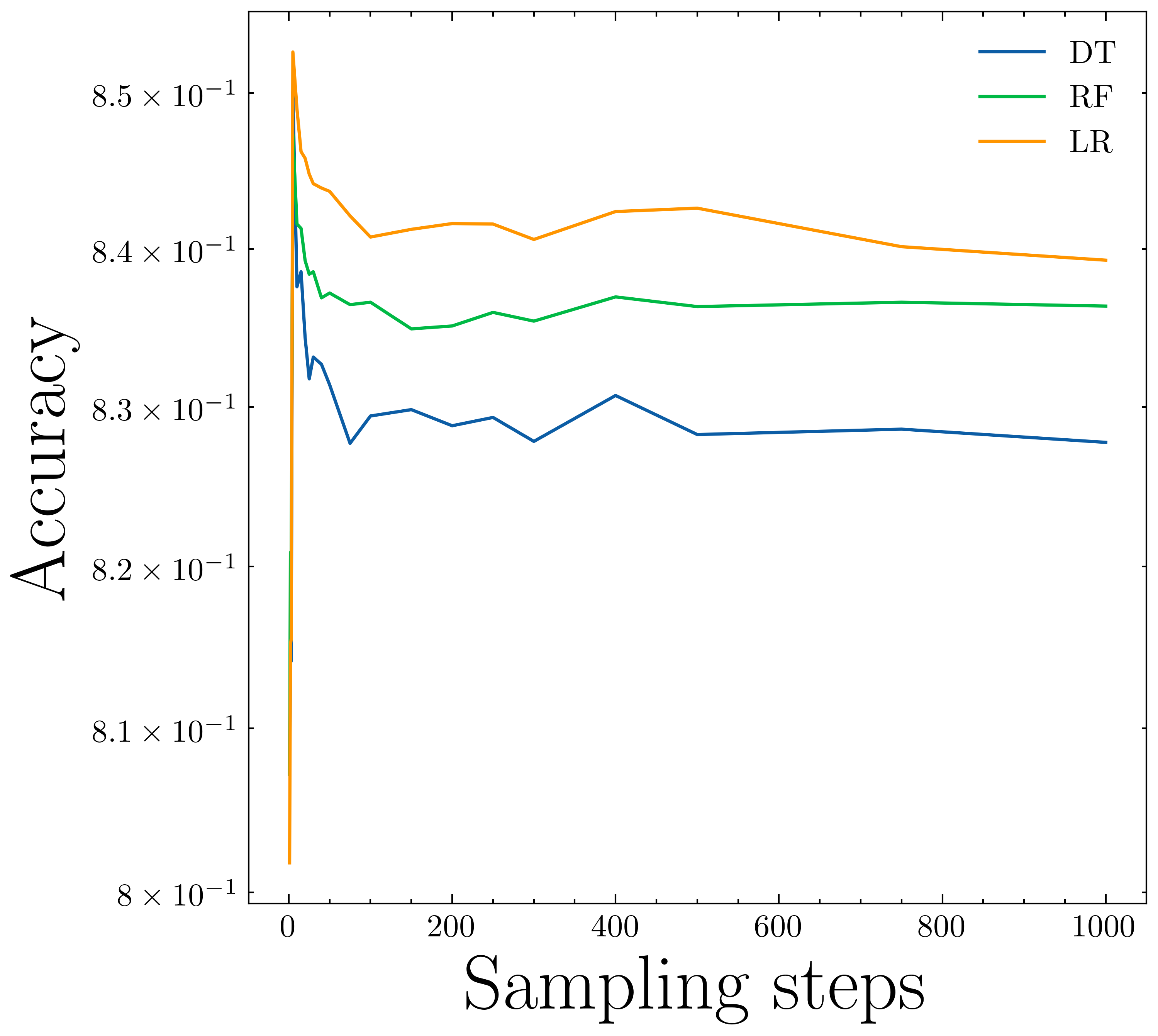}
        \caption{Adult Income}
        \label{fig:adult_steps}
    \end{subfigure}
    
    \vspace{0.5cm}
    
    \begin{subfigure}[b]{0.4\textwidth}
        \centering
        \includegraphics[width=\linewidth]{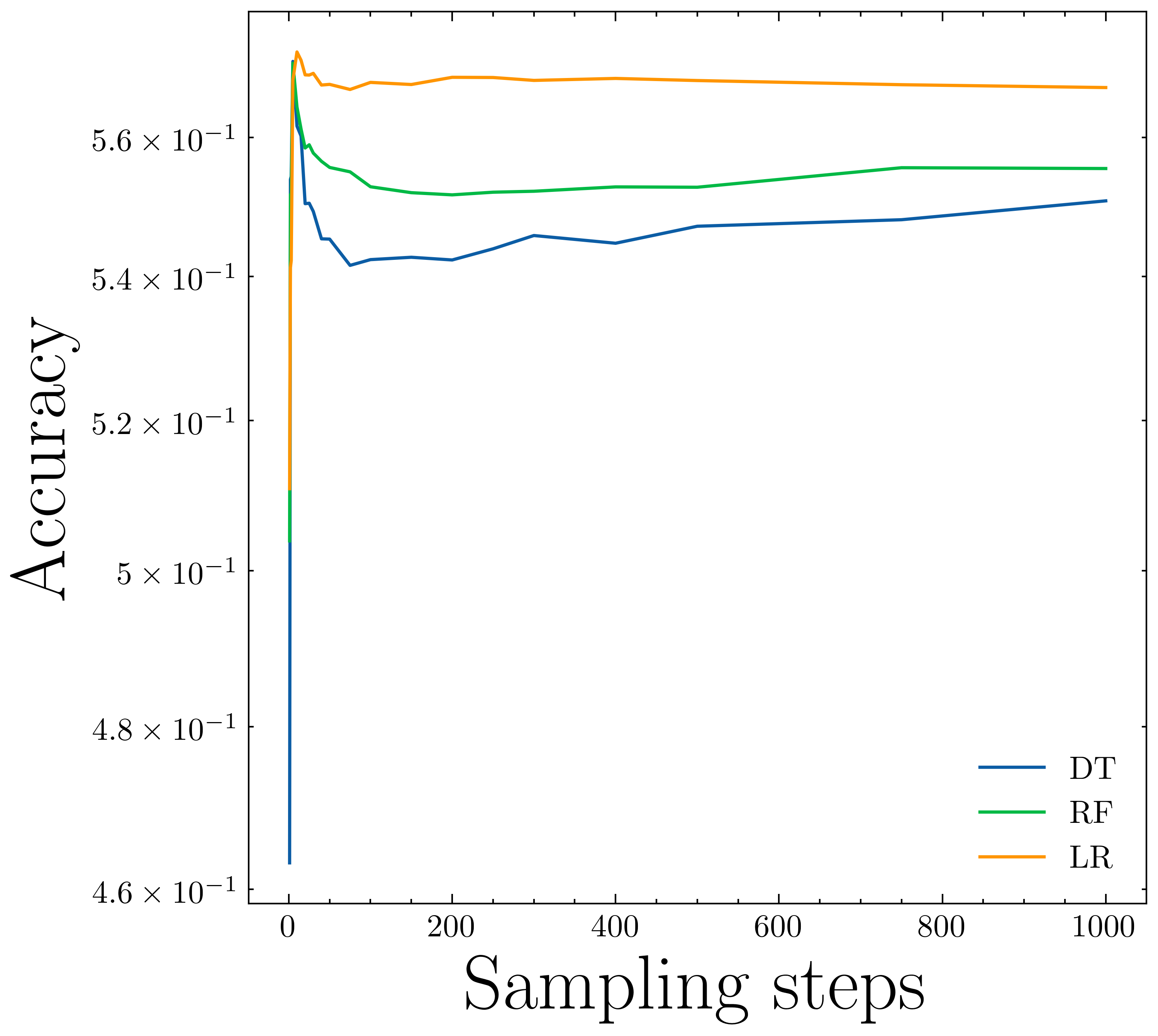}
        \caption{Diabetes}
        \label{fig:diabetes_steps}
    \end{subfigure}
    \hfill
    \begin{subfigure}[b]{0.4\textwidth}
        \centering
        \includegraphics[width=\linewidth]{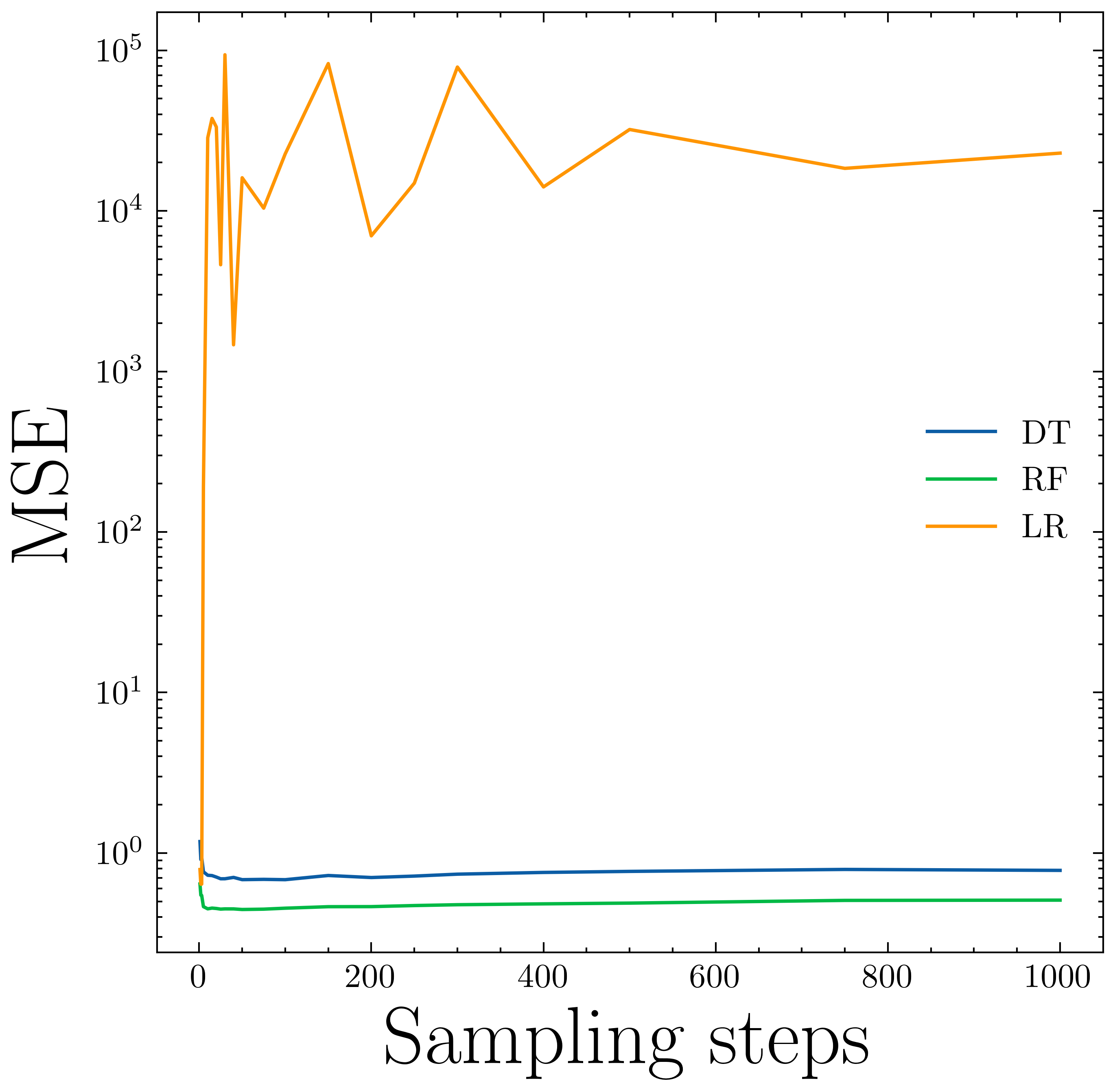}
        \caption{California Housing}
        \label{fig:housing_steps}
    \end{subfigure}

    \caption{Analysis of model performance for different numbers of sampling steps. DT stands for Decision Tree model, RF stands for Random Forest model and LR stands for Linear/Logistic regression model. }
    \label{fig:steps_performance}
\end{figure}

We empirically observed that model performance, measured by accuracy for classification tasks and mean squared error (MSE) for regression tasks, deteriorates as the number of sampling steps increases. Notably, for regression tasks, linear regression models show significantly poorer performance with an increasing number of sampling steps. For our analysis, we utilized an Exponential Moving Average (EMA) denoiser with a guidance scale of 5. Across all datasets, the optimal results were consistently achieved when the number of sampling steps was 5. The relationship between the number of sampling steps and model performance is illustrated in Figure \ref{fig:steps_performance}.

\end{document}